\documentclass[runningheads]{llncs}
\usepackage{graphicx}
\usepackage{amsmath,amssymb} 
\usepackage{color}
\usepackage[width=122mm,left=12mm,paperwidth=146mm,height=193mm,top=12mm,paperheight=217mm]{geometry}

\usepackage{marvosym} 
\usepackage{microtype} 

\usepackage{tabularx}
\usepackage{booktabs}
\usepackage{rotating}

\usepackage[tight]{subfigure}

\usepackage{multirow}
\usepackage{footnote}

\usepackage{caption}

\usepackage{amsmath}
\usepackage{amssymb}
\usepackage{mathrsfs} 
\usepackage{blindtext}

\newcommand{\myparagraph}[1]{\vspace{0.1em}\noindent\textbf{#1}}

\begin{document}
\pagestyle{headings}
\mainmatter

\title{Multi-Person Tracking by Multicut\\and Deep Matching} 

\author{Siyu Tang \qquad Bjoern Andres \qquad Mykhaylo Andriluka \qquad Bernt Schiele}
\institute{Max Planck Institute for Informatics, Saarbr\"ucken, Germany}

\maketitle

\begin{abstract}
  In \cite{tang2015subgraph}, we proposed a graph-based formulation that links and clusters person hypotheses 
  over time by solving a minimum cost subgraph multicut problem.
  In this paper, we modify and extend \cite{tang2015subgraph} in three ways: 
  {\it 1)} We introduce a novel local pairwise feature based on local appearance matching that is robust to partial occlusion and camera motion.
  {\it 2)} We perform extensive experiments to compare different pairwise potentials and to analyze the robustness of the tracking formulation. 
  {\it 3)} We consider a plain multicut problem and remove outlying clusters from its solution. 
  This allows us to employ an efficient primal feasible optimization algorithm that is not applicable to the subgraph multicut problem of \cite{tang2015subgraph}.
  Unlike the branch-and-cut algorithm used there, this efficient algorithm used here is applicable to long videos and many detections.
  Together with the novel feature, it eliminates the need for the intermediate tracklet representation of \cite{tang2015subgraph}. 
  We demonstrate the effectiveness of our overall approach on the MOT16 benchmark \cite{MilanL0RS16}, achieving state-of-art performance.
\end{abstract}

\section{Introduction}

Multi person tracking is a problem studied intensively in computer vision.
While continuous progress has been made, false positive detections, 
long-term occlusions and camera motion remain challenging, 
especially for people tracking in crowded scenes. 
Tracking-by-detection is commonly used for multi person tracking
where a state-of-the-art person detector is employed to generate detection hypotheses for 
a video sequence. In this case tracking essentially reduces to an association task between
detection hypotheses across video frames. 
%
This detection association task is often formulated as an optimization problem with respect to a graph:
every detection is represented by a node; edges connect detections across time frames. 
The most commonly employed algorithms aim to find disjoint paths in such a graph \cite{Pirsiavash:2011:GOG,Segal_2013_ICCV,Andriluka:2008:PTD,Zhang:2008:GDA}. 
The feasible solutions of such problems are sets of disjoint paths which do not branch or merge.
While being intuitive, such formulations cannot handle the multiple plausible detections per person, 
which are generated from typical person detectors.
Therefore, pre- and/or post-processing such as non maximum suppression (NMS) on the detections and/or the final tracks is performed, which often requires careful fine-tuning of parameters.

\begin{figure*}[t]
\centering
      \includegraphics[width=110mm]{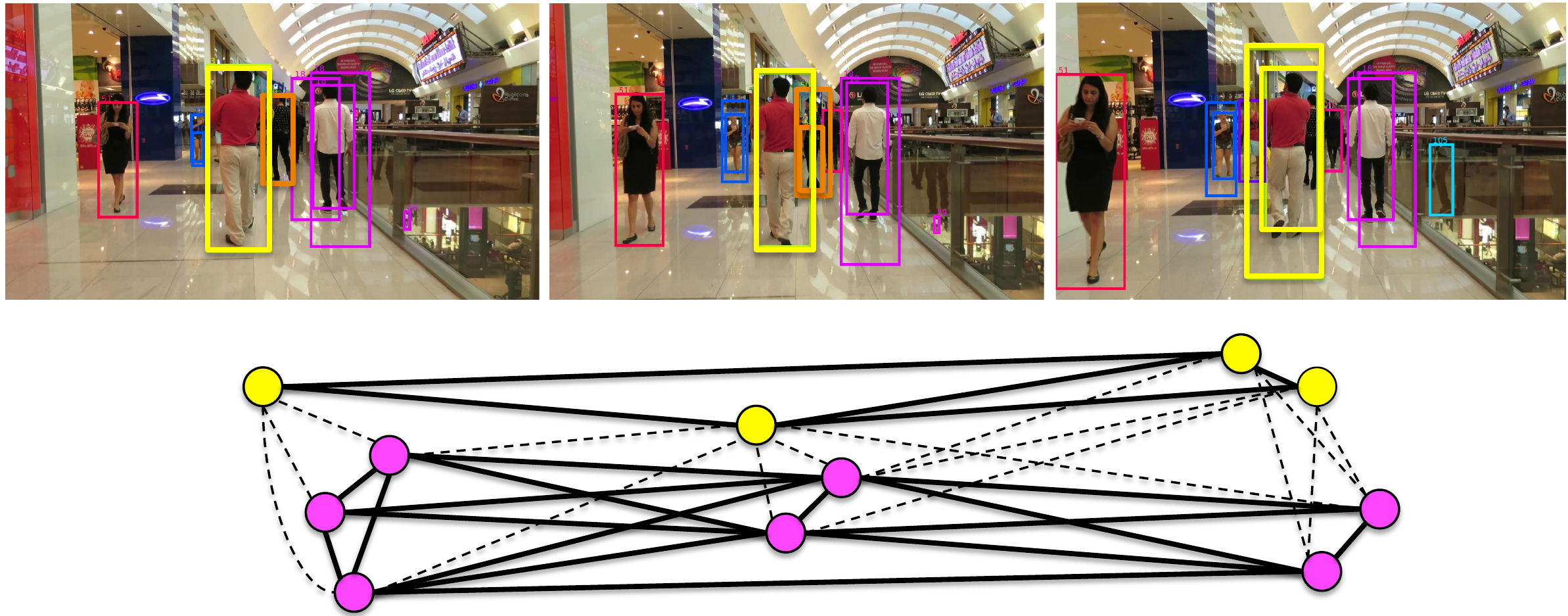}
\caption{An example for tracking by multicut. A graph (bottom) is built based on the detections in three frames (top).
The connected components that are obtained by solving the multicut problem indicate the number of tracks (there are two tracks, depicted in yellow and magenta respectively) as well as the membership of every detection.} 
\label{fig:example}
\end{figure*}

The minimum cost subgraph multicut problem proposed in \cite{tang2015subgraph} 
is an abstraction of the tracking problem that differs conceptually from disjoint path methods.
It has two main advantages:
{\it 1)} Instead of finding a path for each person in the graph, 
it links and clusters multiple plausible person hypotheses (detections) jointly over time and space.
The feasible solutions of this formulation are components of the graph instead of paths.
All detections that correspond to the same person are clustered jointly within and across frames.
No NMS is required, neither on the level of detections nor on the level of tracks.
{\it 2)} For the multicut formulation, 
the costs assigned to edges can be positive, to encourage the incident nodes to be in the same track, 
or negative, to encourage the incident nodes to be in distinct tracks.
Thus, the number and size of tracks does not need to be specified, constrained or penalized and is instead defined by the solution.
This is fundamentally different also from distance-based clustering approaches, e.g.~\cite{6909563} where the cost of joining two detections is non-negative and thus, a non-uniform prior on the number or size of tracks is required to avoid a trivial solution.
Defining or estimating this prior is a well-known difficulty. 
We illustrate these advantages in the example depicted in Fig.~\ref{fig:example}:
We build a graph based on the detections on three consecutive frames, where detection hypotheses within and between frames are all connected.
The costs assigned to the edges encourage the incident node to be in the same or distinct clusters.
For simplicity, we only visualize the graph built on the detections of two persons instead of all.
By solving the minimum cost subgraph multicut problem, a multicut of the edges is found (depicted as dotted lines).
It partitions the graph into distinct components (depicted in yellow and magenta, resp.), each representing one person's track.
Note that multiple plausible detections of the same person are clustered jointly, within and across frames.

\enlargethispage{0.5ex} 
The effectiveness of the multicut formulation for the multi person tracking task is driven by different factors: 
computing reliable affinity measures for pairs of detections; 
handling noisy input detections and 
utilizing efficient optimization methods.
In this work, we extend \cite{tang2015subgraph} on those fronts.
First, for a pair of detections, we propose a reliable affinity measure that is based an effective image matching method DeepMatching \cite{weinzaepfelhal00873592}.
As this 
method matches appearance of local image regions, it is robust to camera motion and partial occlusion. 
In contrast, the pairwise feature proposed in \cite{tang2015subgraph} relies heavily on the spatio-temporal relations of tracklets (a short-term tracklet is used to estimate the speed of a person) which works well only for a static camera and when people walk with constant speed. 
By introducing the DeepMatching pairwise feature, we make the multicut formulation applicable to more general moving-camera videos with arbitrary motion of persons.
Secondly, we eliminate the unary variables which are introduced in \cite{tang2015subgraph} to integrate the detection confidence into the multicut formulation. 
By doing so, we simplify the optimization problem and make it amenable to the fast Kernighan-Lin-type algorithm of \cite{keuper-2015a}.
The efficiency of this algorithm eliminates the need for an intermediate tracklet representation, which greatly simplifies the tracking pipeline.
Thirdly, we integrate the detection confidence into the pairwise terms such that detections with low confidence simply have a low probability to be clustered with any other detection, most likely ending up as singletons that we remove in a post-processing step.
With the above mentioned extensions, we are able to achieve competitive performance on the challenging MOT16 benchmark.

\section{Related Work}
Recent work on multi-person tracking primarily focuses on tracking-by-detection. Tracking
operates either by directly linking people detections over time \cite{kim_ICCV2015_MHTR,Choi15}, or
by first grouping detections into tracklets and then combining those into tracks
\cite{Wang2015arxiv}. A number of approaches rely on data association methods
such as the Hungarian algorithm \cite{xiang2015learning,Bewley16arxiv}, network flow optimization
\cite{Zhang:2008:GDA,Zamir:2012:GMC,Wang2015arxiv,li2009learning}, and multiple hypotheses tracking
\cite{kim_ICCV2015_MHTR}, and combine them with novel ways to learn the appearance of
tracked targets.
%
%
\cite{kim_ICCV2015_MHTR} proposed to estimate a target-specific appearance model online and used a
generic CNN representation to represent person
appearance. In \cite{xiang2015learning} it is proposed to formulate tracking as a Markov decision process
with a policy estimated on the labeled training data. \cite{yang2015temporal} proposes novel appearance
representations that rely on the temporal evolution in appearance of the tracked
target. 
In this paper we propose a pairwise feature that similarly to \cite{Choi_2015_ICCV} is based on
local image patch matching. Our model is inspired by \cite{weinzaepfelhal00873592}
and it operates on pairs of hypotheses which
allows to directly utilize its output as costs of edges on the hypothesis graph. Our pairwise potential
is particularly suitable to our tracking formulation that finds tracks by optimizing a global
objective function. This is in contrast to target-specific appearance methods that are trained
online and require iterative assembly of tracks over time, which precludes globally solving for all
trajectories in an image sequence.

Perhaps closest to our work are methods that aim to recover people tracks by
optimizing a global objective function \cite{Zamir:2012:GMC,Milan:2014:CEM,tang2015subgraph}. 
\cite{Milan:2014:CEM} proposes a continuous
formulation that analytically models effects such as mutual occlusions, dynamics and trajectory
continuity, but utilizes a simple color appearance model. \cite{Zamir:2012:GMC} finds tracks by
solving instances of a generalized minimum clique problem, but due to model complexity resorts to
a greedy iterative optimization scheme that finds one track at a time whereas we jointly recover
solutions for all tracks. We build on the multi-cut formulation proposed in \cite{tang2015subgraph}
and generalize it to large scale sequences based on the extensions discussed below.

\section{Multi-Person Tracking as a Multicut Problem}
\label{section:problem-formulation}


In Section~\ref{section:mc}, we recall the minimum cost multicut problem that we employ as a mathematical abstraction for multi person tracking.
We emphasize differences compared to the minimum cost subgraph multicut problem proposed in \cite{tang2015subgraph}.
In Section~\ref{section:pairwise}, we define the novel DeepMatching feature and its incorporation into the objective function.
In Section~\ref{ClustersToTracks}, we present implementation details.

\subsection{Minimum Cost Multicut Problem}
\label{section:mc}

In this work, multi person tracking is cast as a minimum cost multicut problem 
\cite{chopra-1993}
w.r.t.~a graph $G = (V, E)$ whose node $V$ are a finite set of \emph{detections}, i.e., bounding boxes that possibly identify people in a video sequence.
Edges within and across frames connect detections that possibly identify the same person.
For every edge $vw \in E$, a cost or reward $c_{vw} \in \mathbb{R}$ is to be payed if and only if the detections $v$ and $w$ are assigned to distinct tracks. 
Multi person tracking is then cast as an the binary linear program
\begin{align}
\min_{x \in \{0,1\}^E} \quad
	& \sum_{e \in E} c_e x_e \label{eq:mc-objective} \\
\textnormal{subject to} \quad
	& \forall C \in \textnormal{cycles}(G)\ \forall e \in C:\ x_e \leq \sum_{e' \in C \setminus \{e\}} x_{e'}
\enspace .
	\label{eq:mc-cycle}
\end{align}
Note that the costs $c_e$ can be both positive or negative.
For detections $v,w \in V$ connected by an edge $e = \{v,w\}$, the assignment $x_e = 0$ indicates that $v$ and $w$ belong to the same track. 
Thus, the constraints \eqref{eq:mc-cycle} can be understood as follows: 
If, for any neighboring nodes $v$ and $w$, there exists a path in $G$ from $v$ to $w$ along which all edges are labeled 0 (indicating that $v$ and $w$ belong to the same track), then the edge $vw$ cannot be labeled 1 (which would indicate the opposite).
In fact, \eqref{eq:mc-cycle} are generalized transitivity constraints which guarantee that a feasible solution $x$ well-defines a decomposition of the graph $G$ into tracks. 

We construct the graph $G$ such that edges connect detections not only between neighboring frames but also across longer distances in time.
Such edges $vw \in E$ allow to assign the detections $v$ and $w$ to the same track even if there would otherwise not exist a $vw$-path of detections, one in each frame. 
This is essential for tracking people correctly in the presence of occlusion and missing detections.

\textbf{Differences compared to \cite{tang2015subgraph}}. 
The minimum cost multicut problem
\eqref{eq:mc-objective}--\eqref{eq:mc-cycle},
we consider here differs from
the minimum cost subgraph multicut problem of \cite{tang2015subgraph}.
In order to handle false positive detections, \cite{tang2015subgraph} introduces additional binary variables at the nodes, switching detections on or off.
A cost of switching a dectection on is defined w.r.t.~a confidence score of that detection.
Here, we do not consider binary variables at nodes and incorporate a detection confidence into the costs of edges.
In order to remove false positive detections, we remove small clusters from the solution in a post-processing step.
A major advantage of this modification is that our minimum cost multicut problem 
\eqref{eq:mc-objective}--\eqref{eq:mc-cycle},
unlike the minimum cost subgraph multicut problem of 
\cite{tang2015subgraph},
is amenable to efficient approximate optimization by means of the KLj-algorithm of \cite{keuper-2015a}, without any modification.
This algorithm, unlike the branch-and-cut algorithm of
\cite{tang2015subgraph},
can be applied in practice directly to the graph of detections defined above, thus eliminating the need for the smaller intermediate representation of \cite{tang2015subgraph} by tracklets.

\textbf{Optimization}.
Here, we solve instances of the minimum cost multicut problem approximatively
by means of the algorithm KLj of \cite{keuper-2015a}.
This algorithm iteratively updates bipartitions of a subgraph. 
The worst-case time complexity of any such update is $O(|V||E|)$. 
The number of updates is not known to be polynomially bounded but is small in practice (less than 30 in our experiments). 
Moreover, the bound $O(|V||E|)$ is almost never attained in practice, 
as shown by the more detailed analysis in \cite{keuper-2015a}.

\subsection{Deep Matching based Pairwise Costs}
\label{section:pairwise}

\begin{figure*}
\centering
       \begin{tabular}{cc cc}
	  \includegraphics[height=45mm]{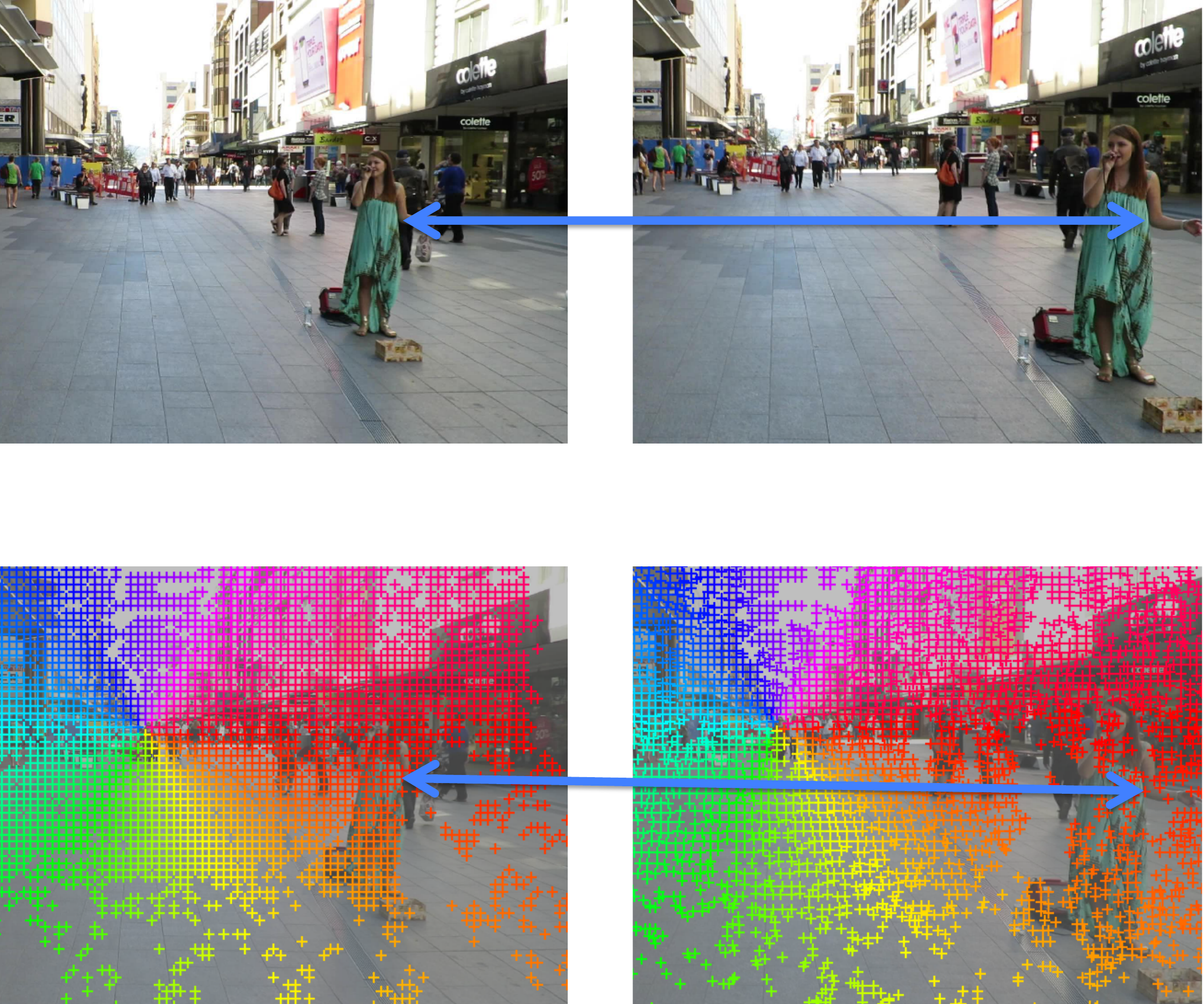} \hspace{2em}
	  &\includegraphics[height=45mm]{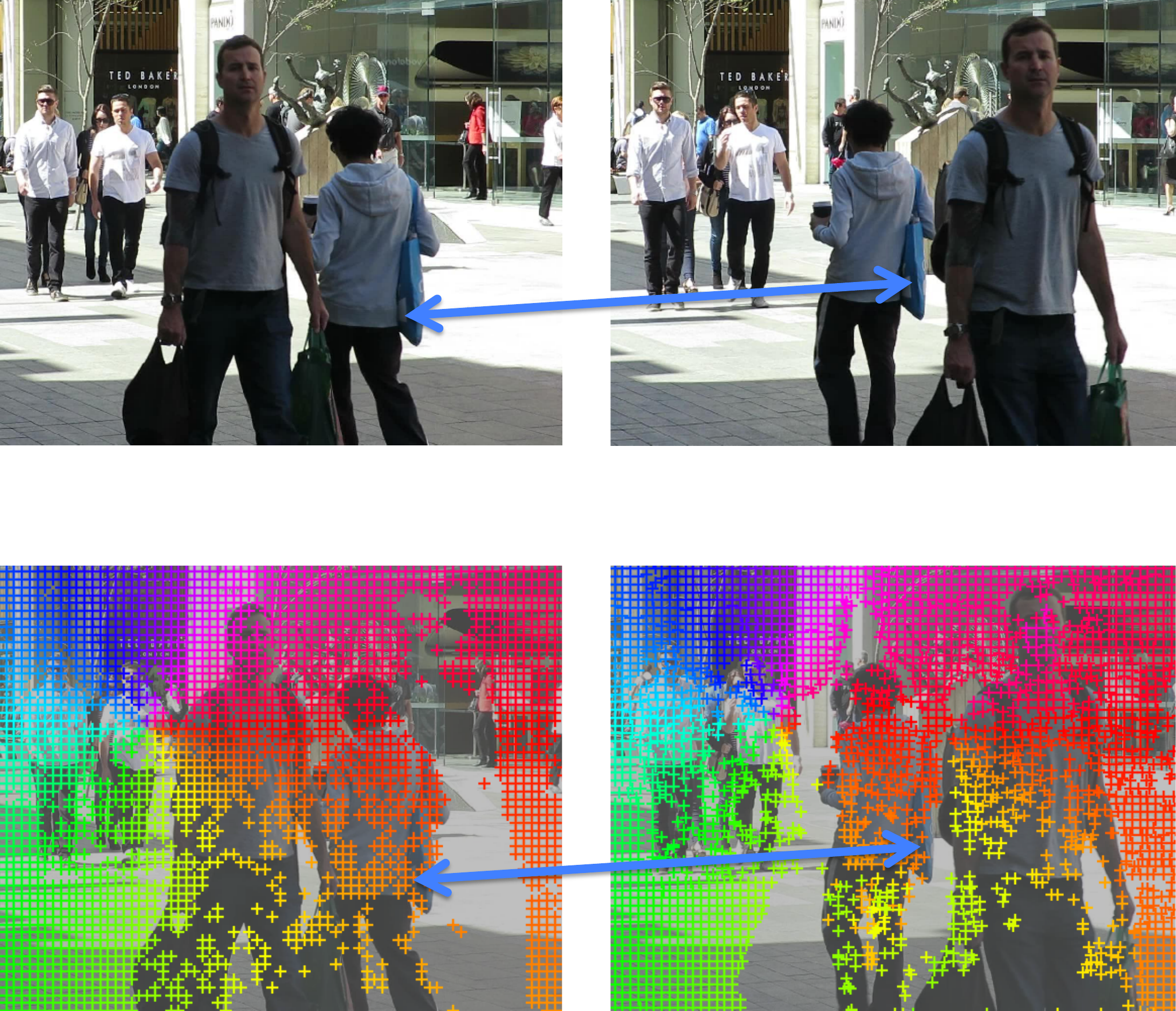} 
	\end{tabular}

	\caption{Visualization of the DeepMatching results on the MOT16 sequences }
		\label{fig:dm-example}
\end{figure*}

In order to specify the costs of the optimization problem introduced above for tracking, we need to define, for any pair of detection bounding boxes, a cost or reward to be payed if these bounding boxes are assigned to the same person.
For that, we wish to quantify how likely it is that a pair of bounding boxes identify the same person.
In \cite{tang2015subgraph}, this is done w.r.t.~an estimation of velocity that requires an intermediate tracklet representation and is not robust to camera motion.
Here, we define these costs exclusively w.r.t.~image content.
More specifically, we build on the significant improvements in image matching made by DeepMatching \cite{weinzaepfelhal00873592}.

DeepMatching applies a multi-layer deep convolutional architecture to yield possibly non-rigid matchings between a pair of images.
Fig.~\ref{fig:dm-example} shows results of DeepMatching for two pairs of images from the MOT16 sequences\footnote{We use the visualization code provided by the authors of \cite{weinzaepfelhal00873592}}.
The first pair of images is taken by a moving camera; the second pair of images is taken by a static camera.
Between both pairs of images, matched points (blue arrows) relate a person visible in one image to the same person in the second image.

Next, we describe our features defined w.r.t.~a matching of points between a pair of detection bounding boxes. 
Each detection bounding box $v \in V$ has the following properties: its spatio-temporal location $(t_v, x_v, y_v)$, scale $h_v$, detection confidence $\xi_v$ and, finally, a set of keypoints $M_v$ inside $v$.
Given two detection bounding boxes $v$ and $w$ connected by the edge $\{v,w\} = e \in E$, 
we define $MU = |M_v \cup M_w|$ and $MI = |M_v \cap M_w|$ and the five features
\begin{align}
f^{(e)}_1 & := MI/MU \\
f^{(e)}_2 & := \min \{\xi_v, \xi_w\} \\
f^{(e)}_3 & := f^{(e)}_1 f^{(e)}_2 \\
f^{(e)}_4 & := (f^{(e)}_1)^2 \\
f^{(e)}_5 & := (f^{(e)}_2)^2 
\end{align}

Given, for any edge $e = \{v,w\} \in E$ between two detection bounding boxes $v$ and $w$, 
the feature vector $f^{(e)}$ for this pair, 
we learn a probability $p_e \in (0,1)$ of these detection bounding boxes to identify distinct persons.
More specifically, we assume that $p_e$ depends on the features $f^{(e)}$ by a logistic form
\begin{align}
p_e := \frac{1}{1 + \exp(-\langle \theta, f^{(e)} \rangle)}
\end{align}
with parameters $\theta$.
We estimate these parameters from training data by means of logistic regression.
Finally, we define the cost $c_e$ in the objective function
\eqref{eq:mc-objective} 
as 
\begin{align}
c_e := \log\frac{p_e}{1-p_e} = \langle \theta, f^{(e)} \rangle
\enspace .
\end{align}

Two remarks are in order:
Firstly, the feature $f^{(e)}_2$ incorporates the detection confidences of $v$ and $w$ that defined unary costs in 
\cite{tang2015subgraph} 
into the feature $f^{(e)}$ of the pair $\{v,w\}$ here.
Consequently, detections with low confidence will be assigned with low probability to any other detection.
Secondly, the features $f^{(e)}_3, f^{(e)}_4, f^{(e)}_5$ are to learn a non-linear map from features $f^{(e)}_1, f^{(e)}_2$ to edge probabilities by means of linear logistic regression.

\subsection{Implementation Details}
\label{ClustersToTracks}

\textbf{Clusters to tracks.} 
The multicut formulation clusters detections jointly over space and time for each target. 
It is straight-forward to generate tracks from such clusters: 
In each frame, we obtain a representative location $(x, y)$ and scale $h$ by averaging all detections that belong to the same person (cluster). 
A smooth track of the person is thus obtained by connecting these averages across all frames. 
Thanks to the pairwise potential incorporating a detection confidence, low confidence detections typically end up as singletons or in small clusters which are deleted from the final solution.
Specifically, we eliminate all clusters of size less than 5 in all experiments.

\textbf{Maximum temporal connection.} 
Introducing edges that connect detections across longer distance in time is essential to track people in the presence of occlusion. 
However, with the increase of the distance in time, the pairwise feature becomes less reliable. 
Thus, when we construct the graph, it is necessary to set a maximum distance in time. 
In all the experiments, we introduce edges for the detections that are at most 10 frames apart. 
This parameter is based on the experimental analysis on the training sequences and is explained in more detail in Section~\ref{subsection-comparison-pairwise}.

\section{Experiments and Results}
We analyze our approach experimentally and compare to prior work on the MOT16 
Benchmark \cite{MilanL0RS16}. The benchmark includes training and test sets composed of 7 sequences each.
We learn the model parameters for the test sequences based on the corresponding training sequences.
We first conduct an experimental analysis that validates the effectiveness of the DeepMatching based affinity measure in
Sec.~\ref{subsection-comparison-pairwise}. 
In Sec.~\ref{robustness-to-detections} we demonstrate that the multicut formulation is robust to detection noise. 
In Sec.~\ref{subsection-MOT-result} we compare our method with the best published results on the MOT16 Benchmark.

\subsection{Comparison of Pairwise Potentials}
\label{subsection-comparison-pairwise}
\myparagraph{Setup.} In this section we compare the DeepMatching (DM) based pairwise potential with 
a conventional spatio-temporal relation (ST) based pairwise potential. 
More concretely, given two detections $v$ and $w$, each has the following properties: spatio-temporal location $(t, x, y)$, scale $h$, detection confidence $\xi$.
Based on these properties the following auxiliary variables are introduced to capture geometric relations between the bounding boxes:
$\Delta x = \frac{|x_v-x_w|}{\bar{h}}, \Delta y = \frac{|y_v-y_w|}{\bar{h}} ,\Delta h = \frac{|h_v-h_w|}{\bar{h}},y = \frac{|y_v-y_w|}{\bar{h}},IOU = \frac{|B_v \cap B_w|}{|B_v \cup B_w|}
,t = {t_v-t_w}$, 
where $\bar{h} = \frac{(h_v + h_w)}{2}$, $IOU$ is the intersection over union of the two detection bounding box areas
and $\xi_{min}$ is the minimum detection score between  $\xi_v$ and $\xi_w$.
The pairwise feature $f^{(e)}$ for the spatio-temporal relations (ST) is then defined as $(\Delta t, \Delta x, \Delta y, \Delta h,  IOU, \xi_{min})$.
Intuitively, the ST features are able to provide useful information within a short temporal window, because they only model the geometric relations between bounding boxes.
DM is built upon matching of local image features that is reliable for camera motion and partial occlusion in longer temporal window.

We collect test examples from the MOT16-09 and MOT16-10 sequences which are recorded with a static camera and a moving camera respectively.
They can be considered as repretentives of the MOT16 sequences in terms of motion, density and imaging conditions.
The positive (negative) pairs of test examples are the detections that are matched to the same (different) persons' ground truth track over time. 
The negative pairs also include the false positive detections on the background.

\myparagraph{Metric.}
The metric is the verification accuracy, the accuracy or rate of correctly classified pairs.
For a pair of images belong to the same (different) person, 
if the estimated joint probability is larger (smaller) than 0.5, the estimation is considered as correct.
Otherwise, it is a false prediction. 

\myparagraph{Results.} We conduct a comparison between the accuracy of the DM feature 
and the accuracy of the ST feature as a function of distance in time.
It can be seen from Tab.~\ref{tab:pairwise-accuracy} that the 
ST feature achieves comparable accuracy only up to 2 frames distance. 
Its performance deteriorates rapidly for connections at longer time.
In contrast, the DM feature is effective and maintains superior accuracy over time. 
For example on the MOT16-10 sequence which contains rapid camera motion, 
the DM feature improves over the ST feature by a large margin after 10 frames 
and it provides stable affinity measure even at 20 frames distance (accuracy = 0.925).
On the MOT16-09 sequence, the DM feature again shows superior accuracy 
than the ST feature starting from $\bigtriangleup t = 2$.  
However, the accuracy of the DM feature on the MOT16-09 is worse than the one on MOT16-10,
suggesting the quite different statistic among the sequences from the MOT16 benchmark.
As discussed in Sec.~\ref{ClustersToTracks}, 
it is necessary to set a maximum distance in time to exclude unreliable pairwise costs.
Aiming at a unique setting for all sequences, 
we introduce edges for the detections that are maximumly 10 frames apart in the rest experiments of this paper.
\begin{table*} [t]
\centering
\scriptsize
 \begin{tabular}{ c| c |c| c| c |c |c  }
      \toprule
      \multicolumn{7}{c}{MOT16-09: Static camera }\\
      \hline
      \!\!\! Feature   & $\bigtriangleup t =1$  & $\bigtriangleup t = 2$  & $\bigtriangleup t =5$  & $\bigtriangleup t = 10$ & $\bigtriangleup t =15$ & $\bigtriangleup t =20$       \\
      \hline

      \!\!\! ST       &  0.972                             &  0.961                              &0.926                            & 0.856                 &0.807                   &0.781                   \\
      \hline
      \!\!\! DM       &  0.970 (\textcolor{red}{-0.2\%})               & 0.963 (\textcolor{blue}{+0.2\%})                &0.946  (\textcolor{blue}{+2\%})               &0.906 (\textcolor{blue}{+5\%})      &0.867 (\textcolor{blue}{+6\%})       &0.820 (\textcolor{blue}{+3.9\%})                 \\
      \toprule
      \multicolumn{7}{c}{MOT16-10: Moving camera }\\
      \hline
      \!\!\! Feature   & $\bigtriangleup t =1$  & $\bigtriangleup t = 2$  & $\bigtriangleup t =5$  & $\bigtriangleup t = 10$ & $\bigtriangleup t =15$ & $\bigtriangleup t =20$  \\
      \hline

      \!\!\! ST       &  0.985                &  0.977                               & 0.942                                &0.903                                 &0.872                 &0.828                  \\
      \hline
      \!\!\! DM       &  0.985                & 0.984  (\textcolor{blue}{+0.7\%})    &0.975 (\textcolor{blue}{+3.3\%})      &0.957  (\textcolor{blue}{+5.4\%})     &0.939   (\textcolor{blue}{+6.7\%}) &0.925 (\textcolor{blue}{+9.7\%})                  \\
      \bottomrule
    \end{tabular}

\vspace{0.5em}
\caption{Comparison of tracking results based on the DM and the ST feature. The metic is the accuracy or rate of correctly classified pairs on the MOT16-09 and the MOT16-10 sequences.}
\label{tab:pairwise-accuracy}
\end{table*}

\subsection{Robustness to Input Detections}
\label{robustness-to-detections}
Handling noisy detection is a well-known difficulty for tracking algorithms.
To assess the impact of the input detections on the tracking result,
we conduct tracking experiments based on different sets of input detections that are obtained by varying  a minimum detection score threshold ($Score_{min}$).
For example, in Tab.~\ref{tab:robustness}, $Score_{min} = -\infty$ indicates that all the detections are used as tracking input;
whereas  $Score_{min} = 1$ means that only the detections whose score are equal or larger than 1 are considered.
Given the fact that the input detections are obtained from a DPM detector \cite{Felzenszwalb2010PAMI},  
$Score_{min} = -\infty$ and $Score_{min} = 1$ are the two extreme cases, where the recall is maximized for the former one and high precision is obtained for the latter one.

\myparagraph{Metric.}
We evaluate the tracking performance of the multicut model that operates on different sets of input detections.
We use the standard CLEAR MOT metrics. 
For simplicity,  in Tab.~\ref{tab:robustness} we report the Multiple Object
Tracking Accuracy (MOTA) that is a cumulative measure that combines the
number of False Positives (FP), the number of False
Negatives (FN) and the number of ID Switches (IDs).

\myparagraph{Results.} 
On the MOT16-09 sequence, when the minimum detection score threshold ($Score_{min}$) is changed from $0.1$ to $-0.3$, 
the number of detection is largely increased (from $3405$ to $4636$), however the MOTA is only decreased by 1 percent (from $44.1\%$ to $43.1\%$).
Even for the extreme cases, where the detections are either  rather noisy ($Score_{min} = -\infty$) or sparse ($Score_{min} = 1$ ),
the MOTAs are still in the reasonable range. The same results are found on the MOT16-10 sequence as well.
Note that, for all the experiments, we use the same parameters, we delete the clusters whose size is smaller than 5 and no further tracks splitting/merging is performed.

These experiments suggest that the multicut formulation is very robust to the noisy detection input.
This nice property is driven by the fact that the multicut formulation allows us to jointly cluster 
multiple plausible detections that belong to the same target over time and space. 

We also report run time in Tab.~\ref{tab:robustness}. 
The Kernighan-Lin multicut solver provides arguably fast solution for our tracking problem.
E.g. for the problem with more than one million edges, the solution is obtained in 88.34 second.
Detailed run time analysis of KL solver are shown in \cite{keuper-2015a}.
\begin{table*}[t]
\centering
 \begin{tabular}{ c@{\hskip 0.2in} c@{\hskip 0.12in}c@{\hskip 0.12in}c@{\hskip 0.12in}c@{\hskip 0.12in}c@{\hskip 0.12in}c@{\hskip 0.12in}c@{\hskip 0.12in}}
      \toprule
      \multicolumn{8}{c}{MOT16-09} \\
      \hline
      \!\!\! $Score_{\min} $   & $-\infty$  & -0.3 & -0.2 & -0.1 & 0  & 0.1 & 1  \\
      \hline
      \!\!\! $|V|$             & 5377       & 4636   & 4320    & 3985   & 3658  & 3405 & 1713\\
      \!\!\! $|E|$             & 565979     & 422725 & 367998  & 314320 & 265174& 229845& 61440\\
      \!\!\! Run time (s)   & 30.48      & 19.28  & 13.46   & 11.88  & 8.39  & 6.76 & 1.71  \\
      \!\!\! MOTA          & 37.9       & 43.1   & 43.1    & 44.9   & 45.8  & 44.1 & 34.1\\
     \toprule
      \multicolumn{8}{c}{MOT16-10}\\
      \hline
      \!\!\! $Score_{\min} $   & $-\infty$  & -0.3   & -0.2    & -0.1 & 0  & 0.1& 1 \\
      \hline
      \!\!\! $|V|$             & 8769       & 6959   & 6299    &5710 &5221 & 4823 &2349\\
      \!\!\! $|E|$             & 1190074     & 755678 & 621024  &511790 &427847& 365949&88673\\
      \!\!\! Run time (s)    & 88.34     & 39.28  &  30.08  &  21.99&  16.13  &  13.66& 1.94\\
      \!\!\! MOTA            & 26.8       & 32.4   & 34.4    & 34.5 &  34.5 &33.9 &23.3\\
    \bottomrule
    \end{tabular}

    \vspace{0.5em}
\caption{Tracking performance on different sets of input detections. $Score_{\min}$ indicates the minimum detection score threshold.
$|V|$ and $|E|$ are the number of nodes (detections) and edges respectively.}
    \label{tab:robustness}
\end{table*}

\begin{table*}
\begin{tabular}{l c c c  c  c c c  c c c c }
      \toprule
      \!\!\! Method                               & MOTA & MOTP & FAF & MT     & ML     & FP   & FN    & ID Sw & Frag & Hz  & Detector \\
      \midrule
      \!\!\! NOMT\cite{Choi_2015_ICCV}            & 46.4 & 76.6 & 1.6 & 18.3\% & 41.4\% & 9753 & 87565 & 359   & 504  & 2.6 & Public    \\
      \!\!\! MHT \cite{kim_ICCV2015_MHTR}         & 42.8 & 76.4 & 1.2 & 14.6\% & 49.0\% & 7278 & 96607 & 462   & 625  & 0.8 & Public    \\
      \!\!\! CEM \cite{Milan:2014:CEM}         & 33.2 & 75.8 & 1.2 & 7.8\% & 54.4\% & 6837 & 114322 & 642   & 731  & 0.3 & Public  \\
        \!\!\! TBD \cite{Geiger2014PAMI}         & 33.7 & 76.5 & 1.0 & 7.2\% & 54.2\% & 5804 & 112587 & 2418   & 2252  & 1.3 & Public  \\
      \midrule
      \!\!\! Ours                                 & 46.3 & 75.7 & 1.09 & 15.5\% & 39.7\% & 6449 & 90713 & 663   & 1115  & 0.8 & Public    \\
      \bottomrule
    \end{tabular}

    \vspace{0.5em}
    \caption{Tracking Performance on MOT16. }
        \label{tab:mot-result}
\end{table*}

\begin{figure}
  \setlength{\abovecaptionskip}{0pt}
  \centering

  \subfigure[MOT16-06]{\includegraphics[height=20mm]{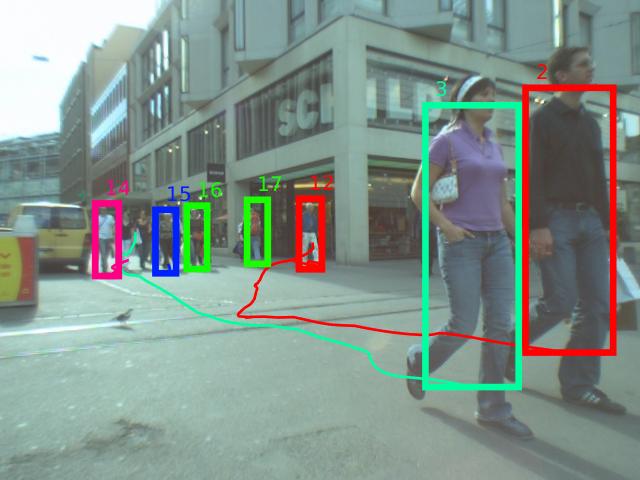}}
  \subfigure[MOT16-12]{\includegraphics[height=20mm]{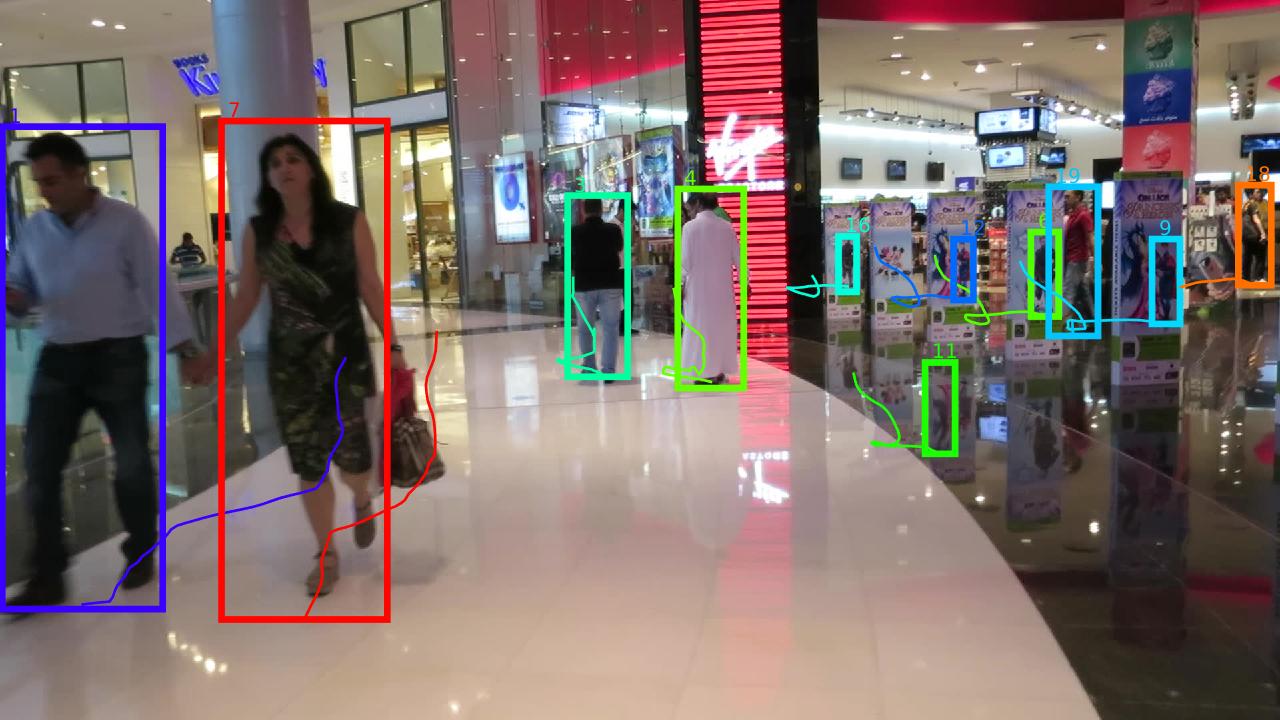}}
  \subfigure[MOT16-03]{\includegraphics[height=20mm]{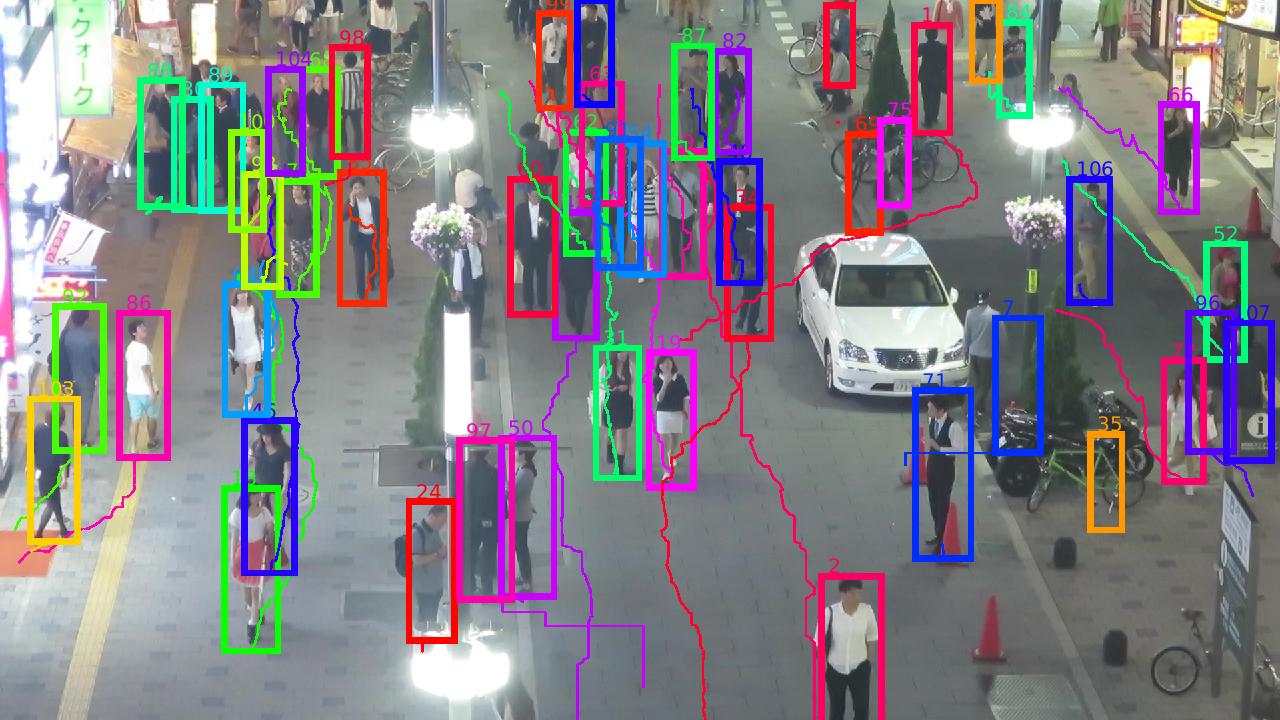}}
  \subfigure[MOT16-08]{\includegraphics[height=20mm]{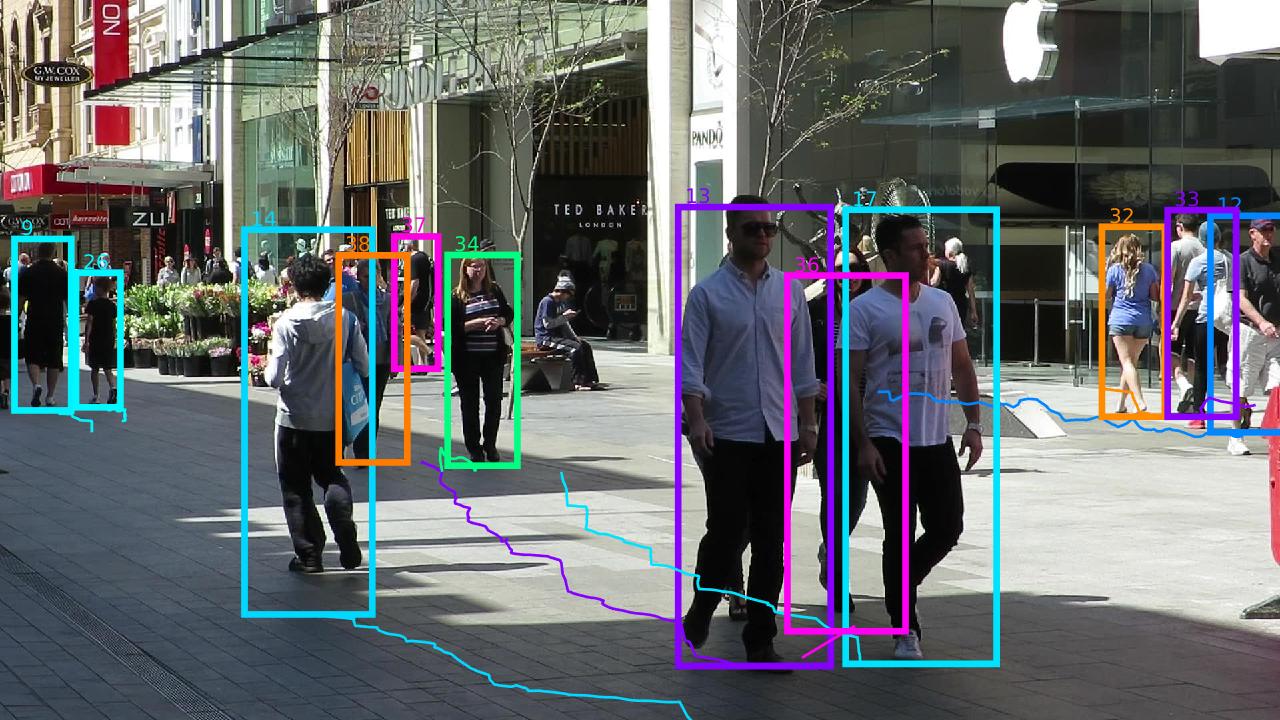}}
  \subfigure[MOT16-07]{\includegraphics[height=20mm]{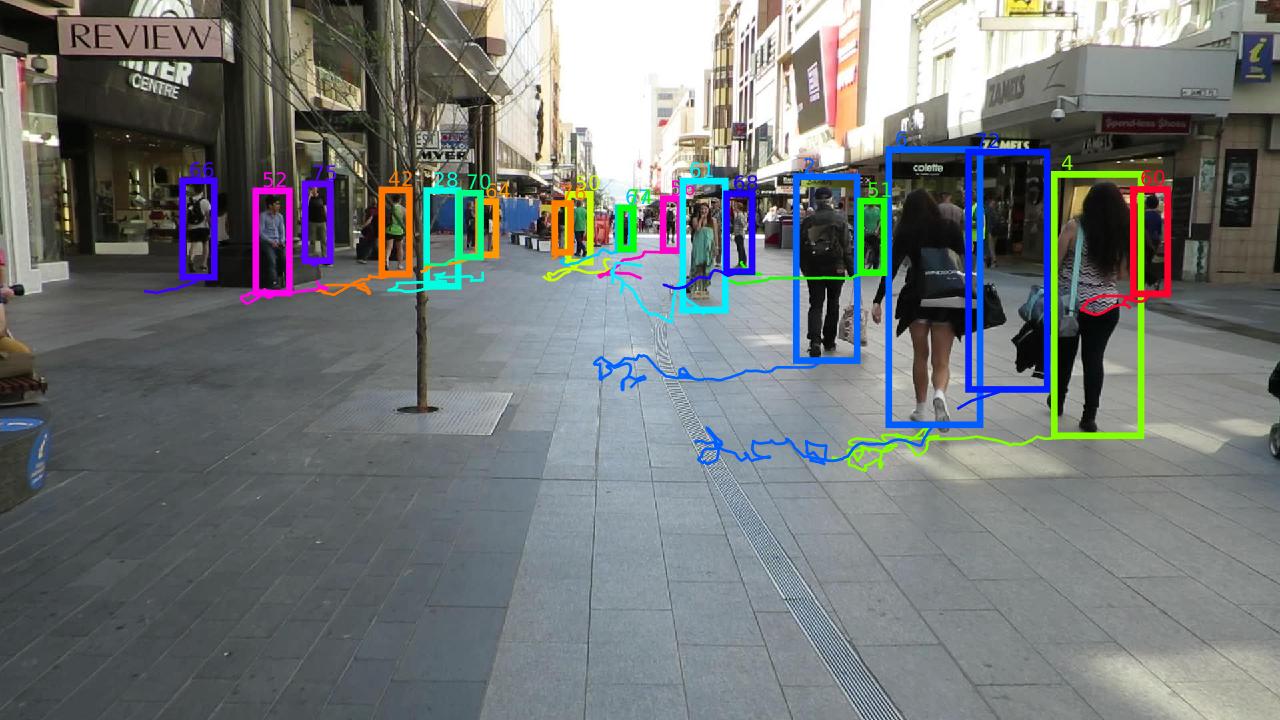}}
  \subfigure[MOT16-01]{\includegraphics[height=20mm]{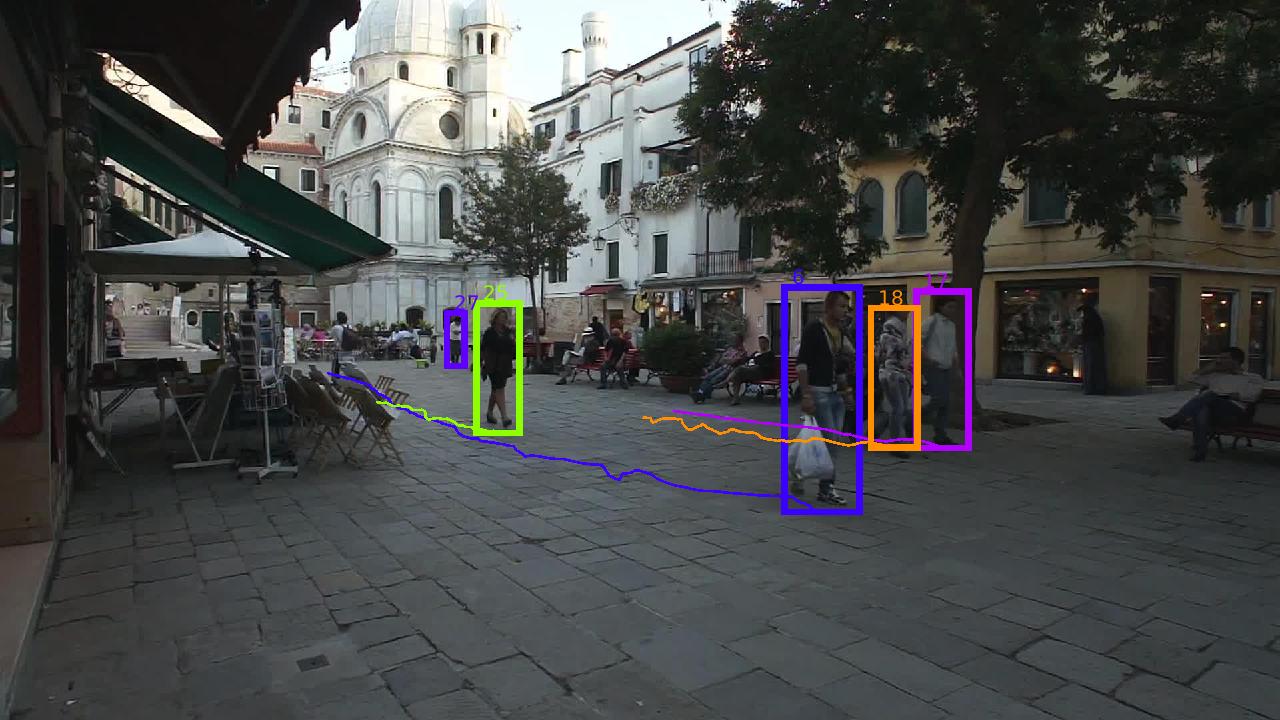}}
  \subfigure[MOT16-09 {\scriptsize (frame 290)}]{\includegraphics[height=20mm]{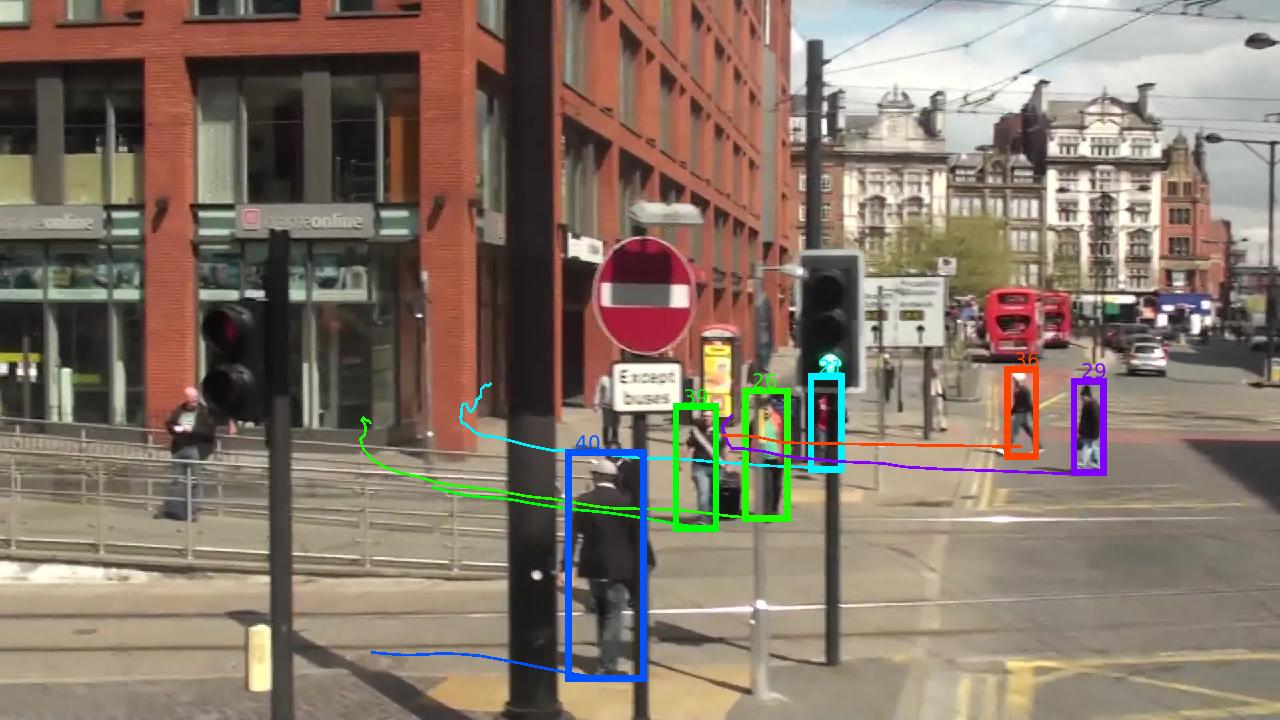}}
  \subfigure[MOT16-09 {\scriptsize (frame 360)}]{\includegraphics[height=20mm]{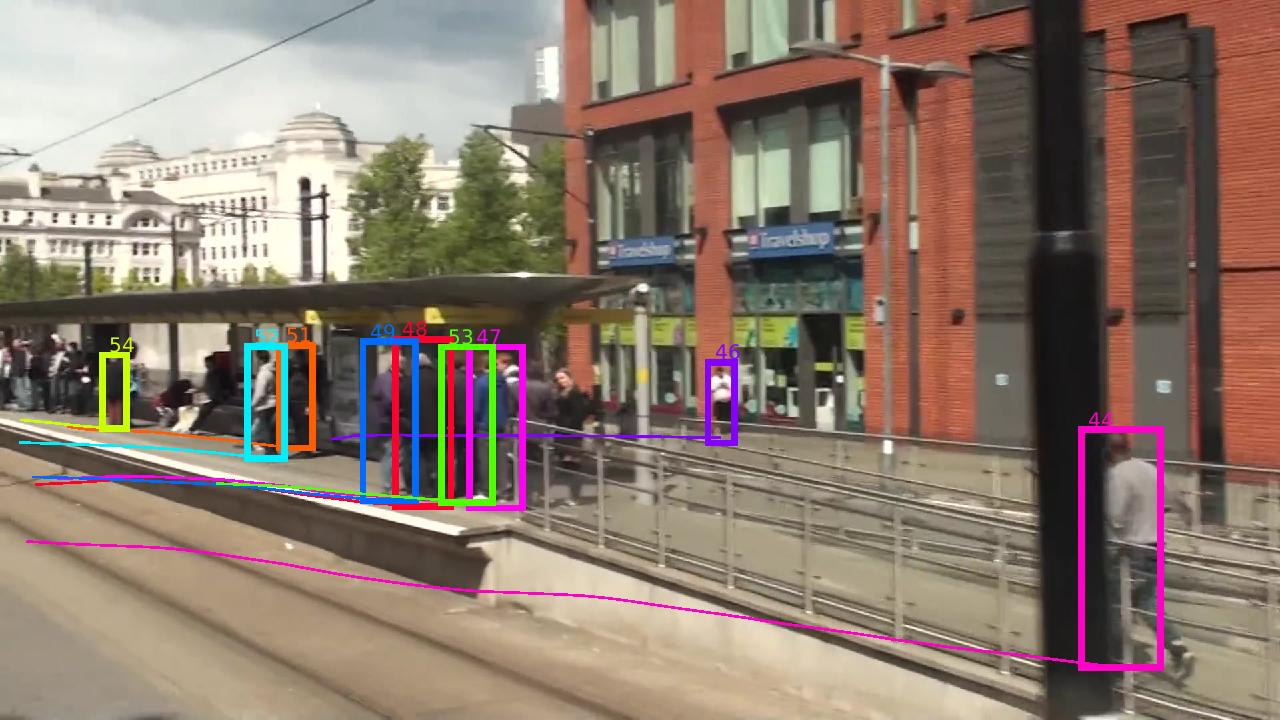}}
  \subfigure[MOT16-09 {\scriptsize (frame 390)}]{\includegraphics[height=20mm]{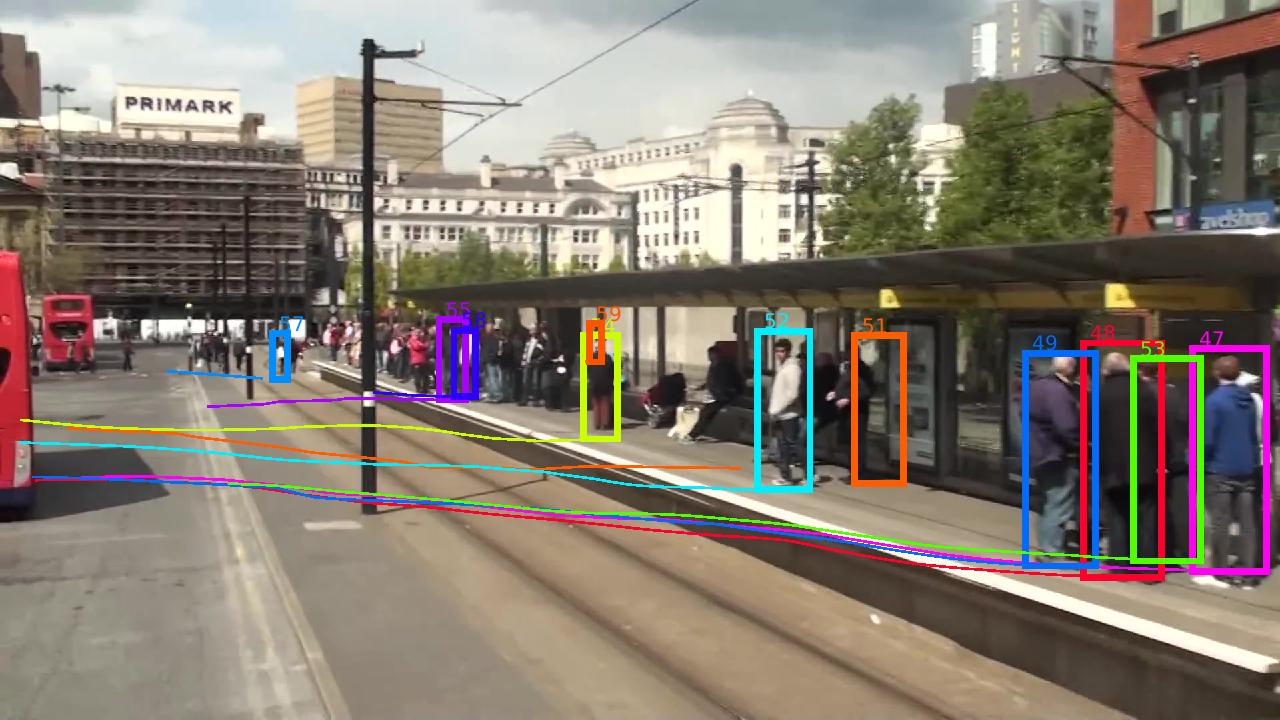}} 
  \caption{Qualitative results for all the sequences from the MOT16 Benchmark.
  The first and second rows are the results from the MOT16-01,MOT16-03,MOT16-06, MOT16-07, MOT16-08 and MOT16-12 sequence.
  The third row is the result from the MOT16-14 sequence when the camera that is carried by a bus is turning fast at  an intersection of two streets.}
\label{fig:mot-example}
\end{figure}
\subsection{Results on MOT16}
\label{subsection-MOT-result}
We test our tracking model on all the MOT16 sequences and submitted our results to the ECCV 2016 MOT Challenge
\footnote{https://motchallenge.net/workshops/bmtt2016/eccvchallenge.html} for evaluation. 
The performance is shown in Tab.~\ref{tab:mot-result}.
The detailed performance and comparison on each sequence will be revealed at the ECCV 2016 MOT Challenge Workshop.
We compare our method with the best reported results including NOMT\cite{Choi_2015_ICCV}, MHT-DAM \cite{kim_ICCV2015_MHTR}, TBD \cite{Geiger2014PAMI}  and CEM \cite{Milan:2014:CEM}.
Overall, we achieve the second best performance in terms of MOTA
with $0.1$ point below the best performed one \cite{Choi_2015_ICCV}.
We visualize our results in Fig.~\ref{fig:mot-example}. 
On the MOT16-12 and MOT16-07 sequences, the camera motion is irregular; whereas on the MOT16-03 and MOT16-08
sequences, scenes are crowded. Despite these challenges,
we are still able to link people through occlusions and produce  long-lived tracks.
The third row of Fig.~\ref{fig:mot-example} show images that are captured by a fast moving camera that is mounted in a bus which is turning at an intersection of two streets.
Under such extreme circumstance, our model is able to track people in a stable and persistent way, demonstrating the reliability of the multicut formulation for multi-person tracking task.

\clearpage

\bibliographystyle{splncs}
\bibliography{short,egbib,bjoern}
\end{document}